\documentclass[conference]{IEEEtran}
\usepackage{times}
\usepackage{graphics}
\usepackage{amssymb}
\usepackage{gradient-text}
\usepackage{tikz}
\usepackage{tabularx}
\usepackage{array}
\usepackage{booktabs}
\usepackage{bm}
\usepackage{algorithm}
\usepackage{algorithmic}
\usepackage{bbm}
\usepackage{adjustbox}
\usepackage{nicematrix}
\usepackage{fontawesome5}
\usepackage[numbers]{natbib}
\usepackage{multicol}
\usepackage[bookmarks=true]{hyperref}
\usepackage{subcaption}
\usepackage{tablefootnote}
\usepackage{flushend}

\definecolor{green}{RGB}{11,155,13}

\newcommand{\tal}{\textsc{tal}~}
\newcommand{\coder}{\textsc{VertiEncoder}}
\newcommand{\former}{\textsc{VertiFormer}}
\newcommand{\vertidecoder}{\textsc{VertiDecoder}}
\newcommand{\encoder}{TransformerEncoder}
\newcommand{\decoder}{TransformerDecoder}
\newcommand{\tr}{Transformer}

\begin{document}

\title{\huge \textsc{\gradientRGB{VertiFormer}{50, 255, 0}{77, 193, 150}}: A Data-Efficient Multi-Task Transformer \\for Off-Road Robot Mobility}

% avoiding spaces at the end of the author lines is not a problem with
% conference papers because we don't use \thanks or \IEEEmembership


\author{\authorblockN{Mohammad Nazeri\authorrefmark{1},
Anuj Pokhrel\authorrefmark{1},
Alexandyr Card\authorrefmark{1},
Aniket Datar\authorrefmark{1},
Garrett Warnell\authorrefmark{2}\authorrefmark{3} and
Xuesu Xiao\authorrefmark{1}}
\authorblockA{\authorrefmark{1}Department of Computer Science,
George Mason University}
% Email: {\tt\scriptsize \{mnazerir, apokhre, acard, adatar, xiao\}@gmu.edu}}
\authorblockA{\authorrefmark{2}DEVCOM Army Research Laboratory}
% Email: {\tt\scriptsize garrett.a.warnell.civ@army.mil}}
\authorblockA{\authorrefmark{3}Department of Computer Science,
The University of Texas at Austin}}

\makeatletter
\g@addto@macro\@maketitle{
  \begin{figure}[H]
  \setlength{\linewidth}{\textwidth}
  \setlength{\hsize}{\textwidth}
  \centering
  \includegraphics[width=1\textwidth]{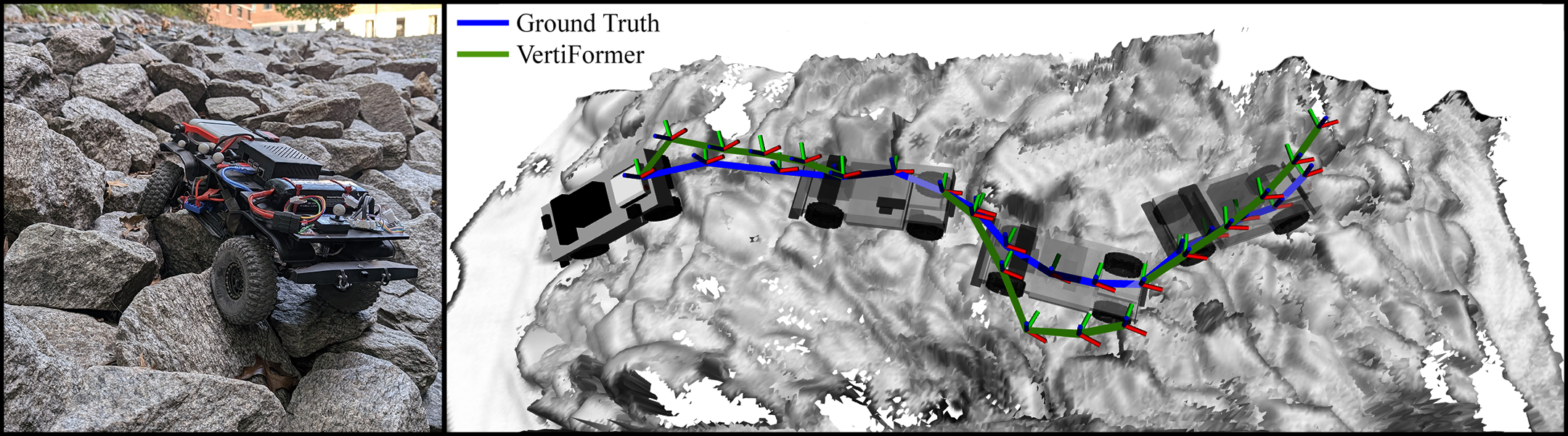}
  \caption{\former~is a data-efficient multi-task Transformer specifically for off-road mobility. Leveraging kinodynamic representation learning, \former~employs unified multi-modal latent representation, learnable masked modeling, and non-autoregressive training to understand complex and nuanced vehicle-terrain interactions with only one hour of training data. }
    \label{fig:cover}
  \end{figure}
}
\makeatother

\maketitle

\addtocounter{figure}{-1}

\begin{abstract}
Sophisticated learning architectures, e.g., Transformers, present a unique opportunity for robots to understand complex vehicle-terrain kinodynamic interactions for off-road mobility. 
While internet-scale data are available for Natural Language Processing (NLP) and Computer Vision (CV) tasks to train Transformers, real-world mobility data are difficult to acquire with physical robots navigating off-road terrain. 
Furthermore, training techniques specifically designed to process text and image data in NLP and CV may not apply to robot mobility. 
In this paper, we propose \former, a novel data-efficient multi-task Transformer model trained with only one hour of data to address such challenges of applying Transformer architectures for robot mobility on extremely rugged, vertically challenging, off-road terrain. 
Specifically, \former~employs a new learnable masked modeling and next token prediction paradigm to predict the next pose, action, and terrain patch to enable a variety of off-road mobility tasks simultaneously, e.g., forward and inverse kinodynamics modeling.  
The non-autoregressive design mitigates computational bottlenecks and error propagation associated with autoregressive models.  
\former's unified modality representation also enhances learning of diverse temporal mappings and state representations, which, combined with multiple objective functions, further improves model generalization. 
Our experiments offer insights into effectively utilizing Transformers for off-road robot mobility with limited data and demonstrate our efficiently trained Transformer can facilitate multiple off-road mobility tasks onboard a physical mobile robot\footnote{\faGithub~\url{https://github.com/mhnazeri/VertiFormer}.}.
\end{abstract}

\IEEEpeerreviewmaketitle

\section{Introduction}
\label{sec:introduction}
Autonomous mobile robots deployed in off-road environments face significant challenges posed by the underlying terrain. 
For example, irregular terrain topographies featuring vertical protrusions from the ground pose extensive risks of vehicle rollover and immobilization~\cite{borges2022survey, lee2023learning, datar2024wheeled}. 
Off-road mobility challenges thus manifest in several critical ways: compromised stability, leading to potential rollover; increased wheel slippage, resulting in reduced traction and impaired locomotion; and the potential for mechanical damage to robots' chassis or drive systems.

Precisely understanding the vehicle-terrain kinodynamic interactions is the key to mitigating such mobility challenges posed by off-road terrain. Although data-driven approaches have shown promises in enabling off-road mobility in relatively flat environments~\cite{overbye2020fast, pan2020imitation, xiao2021learning, sivaprakasam2021improving, fan2021step, karnan2022vi, xiao2022motion, borges2022survey, dashora2022hybrid, triest2022tartandrive, sharma2023ramp, castro2023does, pokhrel2024cahsor, cai2024evora}, the intricate relationships between the robot chassis and vertically challenging terrain, e.g., suspension travel, tire deformation, changing normal and friction forces, and vehicle weight distribution and momentum, motivate the adoption of more sophisticated learning architectures to fully capture and represent the nuanced off-road kinodynamics~\cite{datar2024wheeled}. 

Transformers are the preferred architectures to understand complex relationships, which show promises in Natural Language Processing (NLP)~\cite{radford2018improving, devlin2019bert, radford2019language, brown2020language} and Computer Vision (CV)~\cite{he2022masked, feichtenhofer2022masked, geng2022multimodal, oquab2023dinov2, karypidis2024dinoforesight, patraucean2024trecvit} with self-supervised pre-training emerging as a dominant methodology. 
This trend is now extending to robotics, impacting areas such as manipulation~\cite{o2024open, du2023video, seo2023masked, seo2023multiview, hu2024video} and autonomous driving~\cite{hu2023gaia1, mao2023gptdriver, hu2024drivingworld, bar2024navigation, xiao2024anycar, mattamala2024wild, ai2023invariance}. In addition to the advent of the well-studied \tr~architecture~\cite{vaswani2017attention, dosovitskiy2021image}, this progress is largely attributable to the availability of large-scale datasets~\cite{o2024open, sun2020scalability, nuscenes}
as well as various Transformer training techniques including two primary pre-training paradigms: (i) Masked Modeling (MM) and (ii) autoregressive Next-Token Prediction (NTP)~\cite{chen2024next}. 

However, such benefits are not available nor suitable for off-road robot mobility yet. 
The application of these paradigms to robotics is particularly limited due to the inherent challenges associated with acquiring large-scale robotics datasets, 
especially when outdoor, off-road environments are involved for mobility tasks. Consequently, the effective utilization of data-intensive \tr~models to enable off-road mobility remains an open research question~\cite{firoozi2023foundation}. Further research is also required to investigate the adaptability of existing NLP and CV training paradigms to better suit the unique characteristics of off-road mobility data and tasks.

Motivated by these research gaps, this work presents \former, a novel data-efficient multi-task Transformer model for robot mobility on extremely rugged, vertically challenging, off-road terrain. 
Most notable among all of \former's unique features,  the novel unified latent representation of robot exteroception, proprioception, and action provides a stronger inductive bias and facilitates more effective learning from only one hour of data, compared to the existing practices of separate tokenization of different modalities and sole reliance on the self-attention mechanism to learn inter-modal correlations in NLP and CV with massive datasets. Furthermore, the non-autoregressive nature of \former~avoids error propagation from earlier to later prediction steps and makes \former~faster at inference because it does not require iterative queries for each step.
Additionally, \former's learnable mask enables various off-road mobility tasks within one model simultaneously without the need to retrain separate downstream tasks and mitigates the impact of missing modalities at inference time. 
\former~outperforms the navigation performance achieved by state-of-the-art kinodynamic modeling approaches specifically designed for vertically challenging terrain~\cite{datar2024terrainattentive}, providing empirical evidence supporting the feasibility of training \tr~models on limited robotic datasets using effective training strategies. We also investigate optimal methodologies for employing Transformers, encompassing both \encoder~and \decoder~parts, to facilitate effective learning from limited off-road mobility data. Our contributions can be summarized as follows:

\begin{itemize}
    \item a Transformer architecture, \former, whose unified latent representation, learnable masked modeling, and  non-autoregressive nature simultaneously enable multiple off-road mobility tasks with one hour of data;
    \item a comprehensive evaluation of different Transformer designs, including MM, NTP, Encoder only, and Decoder only, for off-road kinodynamic representation; and
    \item physical on-robot experiments for different off-road mobility tasks on vertically challenging terrain.
\end{itemize}

\section{Related Work}
\label{sec:related_work}
Transformers, initially proposed for language translation task, have demonstrated remarkable versatility across a spectrum of domains, including CV and robotics. This section provides an overview of key advancements in each of these areas, as well as existing work in data-driven off-road mobility. 

\subsection{Transformers in NLP and CV.} 
The \tr~architecture originated from the seminal work of \citet{vaswani2017attention} in machine translation. Subsequent research has explored the effects of different \tr~parts, including using only the \encoder~(BERT~\cite{devlin2019bert}) or \decoder~(GPT series~\cite{radford2018improving, radford2019language, brown2020language}). Other works explored optimization techniques such as adopting a warm-up phase for training Transformers~\cite{xiong2020layer}, specific initialization and optimization methods to train deep Transformers with limited data~\cite{xu2021optimizing}, as well as normalization techniques~\cite{loshchilov2024ngpt}.

Early explorations of Transformers in CV include iGPT~\cite{chen2020generative}. A significant breakthrough came with the introduction of Vision Transformers (ViT) by~\citet{dosovitskiy2021image}. Subsequent research focused on refining training methodologies and enhancing performance, such as incorporating auxiliary tasks~\cite{liu2021efficient} for spatial understanding, two-stage training (self-supervised view prediction followed by supervised label prediction)~\cite{gani2022how}, different token representations~\cite{mao2022discrete}, architectural modifications~\cite{zhai2022scaling}, working in embedding space by JEPA family~\cite{assran2023self, bardes2023mcjepa, bardes2024revisiting}, data augmentation and regularization~\cite{steiner2022how}, and Masked Autoencoders~\cite{he2022masked} with random patch encoding for training stabilization~\cite{chen2021empirical}. Similar to the autoregressive nature of NLP tasks,~\citet{rajasegaran2025empirical} provided empirical guidelines to train Transformers on large-scale video data autoregressively.
Despite the plethora of NLP and CV Transformers trained with internet-scale datasets, existing common training practices may not apply to robot learning with small real-world data, especially for off-road robot mobility. 

\subsection{Transformers in Robotics.}
Recent years have witnessed a surge in the application of Transformers to robotics, encompassing both perception and planning: 
Generalist robot policies based on Transformers, e.g., Octo~\cite{octomodelteam2024octo} and CrossFormer~\cite{doshi2024scaling}, with multi-modal sensory input~\cite{jones2025sight} and action tokenization~\cite{pertsch2025fast} aim to handle diverse tasks such as manipulation and navigation;
Studies in target-driven~\cite{du2021vtnet, wang2024navformer, nazeri2024vanp, huang2024goalguided} and image-goal navigation~\cite{pelluri2024transformers, liu2024citywalker} show that Transformers significantly outperform traditional behavior cloning baselines~\cite{pomerleau1988alvinn, bojarski2016end, nazeri2021exploring}; 
Reinforcement learning has been significantly enhanced by integrating the \tr~architecture, providing improved sequence modeling ~\cite{zhang2024naviformer} and decision-making capabilities~\cite{chen2021decision};  
Transformers have also been used in motion planning to guide long-horizon navigation tasks~\cite{lawson2023control} and reduce the search space for sampling-based motion planners~\cite{johnson2022motion};
In Unmanned Surface Vehicles (USV), MarineFormer~\cite{kazemi2024marineformer} utilizes Transformers to learn the flow dynamics around a USV and then learns a navigation policy resulting in better path length and completion rate.

A common characteristic of these models is their treatment of each sensor modality (e.g., vision, touch, and audio) as a distinct token, relying on the \tr~to learn the inter-modal correlations and their temporal dynamics. While this approach allows for flexible integration of diverse sensory information, it necessitates substantial amounts of training data to compensate for the lack of inductive bias inherent in Transformers~\cite{dosovitskiy2021image}. This data dependency poses a significant challenge, particularly in off-road robot mobility, where real-world, outdoor data acquisition can be expensive and time-consuming. Consequently, there remains a critical need for research focused on refining training methodologies and exploring architectural modifications specifically tailored to address the data scarcity often encountered in robotics.

\subsection{Learning Off-Road Mobility.}
While most learning approaches for off-road autonomy focus on perception tasks~\cite{xiao2022motion, wigness2019rugd, jiang2021rellis}, researchers have recently investigated off-road mobility to account for vehicle stability~\cite{bae2021curriculum, lee2023learning, datar2024learning, pokhrel2024cahsor}, wheel slippage~\cite{siva2019robot, siva2022nauts, sharma2023ramp}, and terrain traversability~\cite{fan2021step, triest2022tartandrive, castro2023does, seo2023learning, cai2024evora}. A relevant work by \citet{xiao2024anycar} aims to use Transformers to enable a universal forward kinodynamics model that can drive different ground vehicles. Most of these approaches adopted specific techniques designed to address one particular off-road mobility task with non-Transformer architectures. 

Focusing on kinodynamic representation for off-road mobility, our non-autoregressive \former~employs a novel variation of MM and NTP paradigms and a unified modality latent representation to predict the next pose, action, and terrain patch in order to simultaneously enable a variety of off-road mobility tasks, e.g., forward and inverse kinodynamics modeling, behavior cloning, and terrain patch reconstruction, without a specific training procedure for each.  

\section{\former}
\label{sec:approach}
We introduce \former, a data-efficient multi-task \tr~model for kinodynamic representation and navigation on complex, vertically challenging, off-road terrain. We propose an efficient training methodology for training \former~utilizing limited (one hour) robotics data, including unified multi-modal latent representation, learnable masking, and non-autoregressive training to improve data efficiency by enabling multi-task learning. 

\begin{figure*}
  \centering
  \includegraphics[width=2\columnwidth]{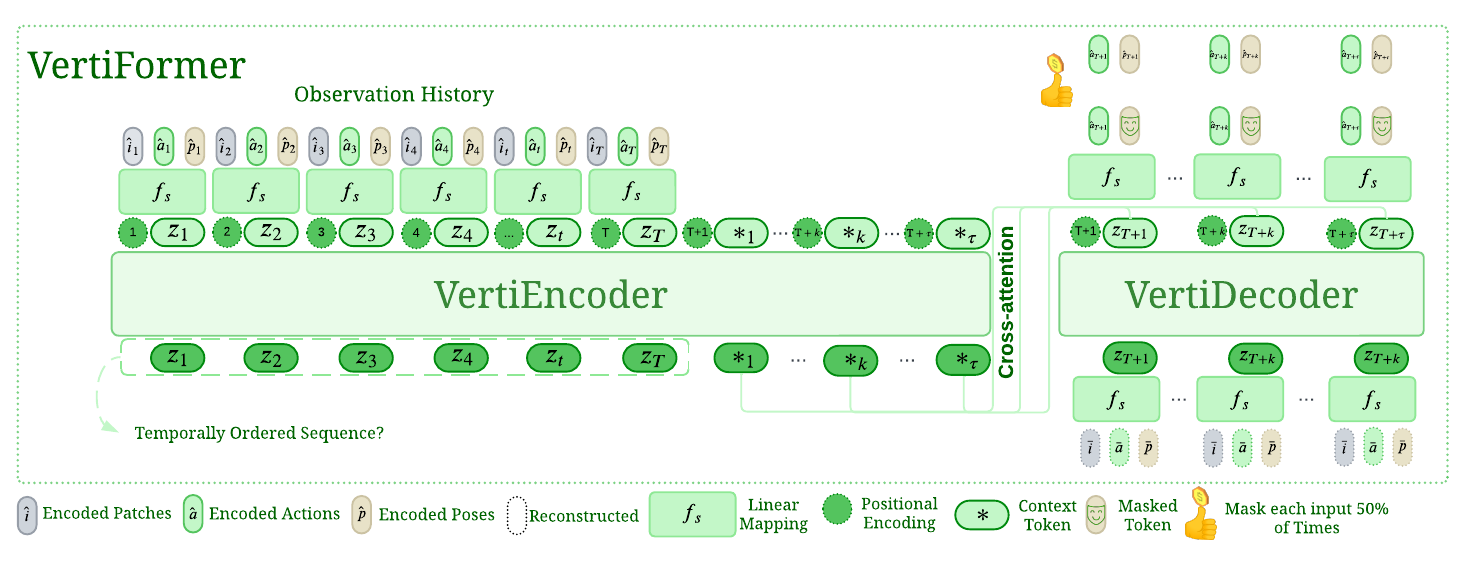}
  \caption{\textbf{\former~Architecture}. \former~employs a TransformerEncoder (left) to receive a history of terrain patches, actions, and poses along with multiple context tokens. To predict future states, the model computes cross-attention between these context tokens and the masked upcoming actions or poses. Causal masking is implemented during this cross-attention computation to ensure that predictions are conditioned only on past and present information, preventing information leakage from future time steps.}
  \label{fig:former}
  % \vspace{-1.2em}
\end{figure*}

\subsection{\former~Training} 
\subsubsection{Unified Multi-Modal Latent Representation}
\label{sec:unified}
\former~consists of both \encoder~(\coder) and \decoder~(\vertidecoder), as illustrated in Fig.~\ref{fig:former} left and right, respectively. Consistent with established practices~\cite{datar2024terrainattentive,nazeri2024vertiencoder}, \former~receives a multi-modal sequence of actions $\mathbf{a_{\text{0:T}}}$, robot poses $\mathbf{p_{\text{0:T}}}$, and the underlying terrain patches $\mathbf{i_{\text{0:T}}}$. The \coder~first applies an independent linear mapping to each modality. Specifically, action commands $\mathbf{a_{\text{0:T}}}$ are projected into an embedding space via a linear function $f_a$, yielding $\mathbf{\hat{a}_{\text{0:T}}}$. Analogously, robot poses $\mathbf{p_{\text{0:T}}}$ and terrain patches $\mathbf{i_{\text{0:T}}}$ are transformed using linear mappings $f_p$ and $f_i$ respectively, producing a sequence of embeddings $\mathbf{\hat{p}_{\text{0:T}}}$ and $\mathbf{\hat{i}_{\text{0:T}}}$. This initial linear mapping can be formally expressed as:
\begin{align}
\hat{a}_{\text{t}} &= f_a(a_{\text{t}}) = W_a a_{\text{t}} + b_a , a_{\text{t}} \in  \mathbf{a_{\text{0:T}}}, \\
\hat{p}_{\text{t}} &= f_p(p_{\text{t}}) = W_p p_{\text{t}} + b_p , p_{\text{t}} \in \mathbf{p_{\text{0:T}}}, \\
\hat{i}_{\text{t}} &= f_i(i_{\text{t}}) = W_i i_{\text{t}} + b_i , i_{\text{t}} \in \mathbf{i_{\text{0:T}}},
\end{align}
where $W_a$, $W_p$, and $W_i$ represent the weight matrices, and $b_a$, $b_p$, and $b_i$ denote the bias vectors for each respective modality.

To facilitate effective cross-modal interaction within \former, it is crucial to establish a consistent distributional characteristic across the modality-specific embeddings. Therefore, a subsequent linear transformation, denoted by 
$f_s$, is applied to the concatenation ($\cdot$) of embeddings:
\begin{align}
z_{\text{t}} &= f_s(\hat{a}_{\text{t}}, \hat{p}_{\text{t}}, \hat{i}_{\text{t}}) = W_s(\hat{a}_{\text{t}} \cdot \hat{p}_{\text{t}} \cdot \hat{i}_{\text{t}}) + b_s , t \in [0:T],
\end{align}
with $W_s$ and $b_s$ denoting the weight matrix and bias vector for $f_s$, respectively. This shared linear mapping $f_s$ aims to project all embeddings into a unified latent space, minimizing potential discrepancies in statistical properties. The resulting unified tokens, $\mathbf{z_{\text{0:T}}}$, are then passed as input to the \coder~(Fig.\ref{fig:former} top left). This procedure ensures a homogeneous input representation for the subsequent encoding layers, crucial for effective multi-modal fusion of robotic data. Empirical results (Fig.~\ref{fig:unified_state}) supporting the importance of such a unified representation, in contrast to the conventional individual modality representations, will be presented in Section~\ref{sec:study}.

\subsubsection{Learnable Masking for Multi-Task Learning}
Combined with our unified representation, we also propose a stochastic learnable MM technique (Fig.\ref{fig:former} top right) to allow \former~to perform multiple predictive tasks, including next pose prediction, action prediction, behavior cloning, and terrain patch prediction (Fig.\ref{fig:former} bottom right). This multi-task learning paradigm is hypothesized to enhance data efficiency by leveraging shared latent representations across related tasks, thereby mitigating the challenges associated with restricted data availability.
During training, we first warm up the model for a few epochs with all modalities, then two distinct data masking methods are applied with equal probability (Fig.\ref{fig:former} top right):
\begin{itemize}
    \item \textbf{Action-Conditioned Pose Prediction:} In 50\% of the training instances, actions generated by human demonstration $\tau$ steps into the future, denoted as $\mathbf{a_{\text{T+1:T}+\tau}}$, are provided as input. Concurrently, the corresponding future poses, $\mathbf{p_{\text{T+1:T}+\tau}}$, are replaced with a learnable mask. This configuration compels the model to predict future poses conditioned on the provided future actions and the preceding historical context, similar to the \emph{Forward Kinodynamic Modeling} (FKD) task in off-road mobility. 

    \item \textbf{Pose-Conditioned Action Prediction:} In the remaining 50\% of instances, the inverse scenario is implemented. Future poses, $\mathbf{p_{\text{T+1:T}+\tau}}$, are provided as input, while the corresponding future actions, $\mathbf{a_{\text{T+1:T}+\tau}}$, are masked using another learnable mask. This prompts the model to predict future actions conditioned on the provided future poses and the historical context, similar to the \emph{Inverse Kinodynamic Modeling} (IKD) task in off-road mobility. 
\end{itemize}
This alternating masking strategy along with our unified representation promotes the learning of a joint representation that is capable of decoding both action and pose information. The utilization of this novel learnable mask allows the model to dynamically adapt the masking pattern. The learnable mask can be conceptualized as a learnable gating mechanism that selectively filters information flow during training.

Furthermore, by extending this masking strategy to mask both future actions, $\mathbf{a_{\text{T+1:T}+\tau}}$, and future poses, $\mathbf{p_{\text{T+1:T}+\tau}}$, simultaneously, \former~is able to perform \emph{Behavior Cloning} (BC) in a zero-shot manner. In this configuration, the model predicts both actions and poses solely based on the historical context, effectively mimicking the demonstrated behavior without requiring explicit information about future actions and poses from a planner.

\subsubsection{Non-Autoregressive Training}
Building upon the works by \citet{octomodelteam2024octo} and \citet{doshi2024scaling}, \former~employs multiple context tokens to represent a distribution of plausible future states. These context tokens serve to inform \vertidecoder~in predicting both the future ego state and the evolution of the environment. Having multiple context tokens allows \former~to predict the future non-autoregressively. The non-autoregressive nature of the proposed architecture is motivated by the potential computational bottlenecks inherent in autoregressive models, which require querying the model multiple times and are subject to drifting due to error propagation from earlier steps. By learning multi-context representations, the non-autoregressive approach aims to improve both training efficiency and inference speed---a critical consideration for real-time robotic control applications.

We train \former~by minimizing the Mean Squared Error (MSE) between the model's predictions and the corresponding ground truth values. Model evaluation is performed by calculating the error rate between the model's predictions and the ground truth values on a held-out, unseen dataset.

\subsection{\former~Inference} 
During FKD inference, \coder~receives the same historical input as training. \vertidecoder~receives sampled actions from an external sampling-based planner (e.g., MPPI~\cite{williams2017model}) while masking the corresponding poses, compelling the model to predict future poses based solely on the sampled actions (and the context tokens) so that the planner can choose the optimal trajectory to minimize a cost function. For IKD, a global planner generates desired future poses, and by masking the actions we encourage the model to predict future actions to achieve these globally planned poses. By masking both actions and poses, \former~can perform zero-shot BC.

As a reference, we examine the average error rate of \former's pose predictions across $\tau=3$ future time steps in one second (3 Hz). We focus on the average error rate across the three pose components, $\mathbf{X}$, $\mathbf{Y}$, and $\mathbf{Z}$. The performance of \former~is compared against two baseline models: \tal~\cite{datar2024terrainattentive} and \citet{nazeri2024vertiencoder}. Notice that \textsc{tal} is a highly accurate forward kinodynamic model specifically designed for vertically challenging terrain, and \citet{nazeri2024vertiencoder} only employs a \encoder~with random masking. 

\begin{table}[h]
    \centering
    % \vspace{0.5em}
    \begin{NiceTabular}{lccc}

    \toprule
    & \tal~\cite{datar2024terrainattentive} & \citet{nazeri2024vertiencoder} & \former \\
    \midrule                       %    TAL   Encoder   Former
    \textbf{{Error Rate}~$\downarrow$} & 0.528 & 0.516    & 0.495 \\
    \bottomrule
    \end{NiceTabular}
    
    \label{tab::reference}
    % \vspace{-2.0em}
\end{table}
We provide \former's architecture parameters in Appendix~\ref{app:architecture} and qualitative samples of FKD in Fig.~\ref{fig:qualitative} of Appendix~\ref{app:qualitative}. The implementation details along with the one-hour dataset description are provided in Appendix~\ref{app:implementation}.

\section{Training \former~with One Hour of Data}
\label{sec:study}
We conduct extensive experiments to demonstrate the efficacy of various features of \former~to allow it to be trained with only one hour of data. We also present our findings in a way that highlights \former’s differences compared to common practices in NLP and CV, where Transformer training practices have been extensively studied~\cite{xiong2020layer, xu2021optimizing, loshchilov2024ngpt, chen2021empirical, liu2021efficient, gani2022how, steiner2022how}. 
Therefore, our experiment results also serve as a guideline on how to optimize Transformer training for robotics, particularly in off-road navigation and mobility tasks with complex vehicle-terrain interactions under data-scarce conditions. 

\former's one hour of training data comes from human-teleoperated demonstration of driving an open-source four-wheeled ground vehicle~\cite{datar2024wheeled} on a custom-built off-road testbed composed of hundreds of rocks and boulders. The demonstrator mostly aims to drive the robot to safely and stably traverse the vertically challenging terrain, but still occasionally encounters dangerous situations such as large roll angles and getting stuck between rocks. Fortunately, those situations serve as explorations for \former~to understand a wider range of kinodynamic interactions. 
Direct application of standard Transformer training methodologies in NLP and CV to such a small robotics dataset proves challenging due to the inherent lack of inductive bias in Transformers~\cite{dosovitskiy2021image}, which necessitates substantial amounts of data for effective training. However, our experiments suggest that \former's judicious modifications to established MM and NTP training paradigms can facilitate effective Transformer training even with limited robotics data. 

We conduct our experiments based on three perspectives: Section~\ref{sec:basic_perspective} provides an analysis of basic factors to train Transformers in general; Section~\ref{sec:robotic_perspective} analyzes the best practices to train Transformers when dealing with off-road robot mobility data; Finally, Sec.~\ref{sec:objective_perspective} evaluates the effectiveness of each off-road mobility learning objective and compares \encoder, \decoder, and non-Transformer end-to-end model performances. For fairness, all experiments are conducted with the same hyper-parameters. 

\subsection{Experiment Results of Basic Transformer Factors} \label{sec:basic_perspective}

\begin{figure}[t]
  \centering
  \includegraphics[width=0.7\columnwidth]{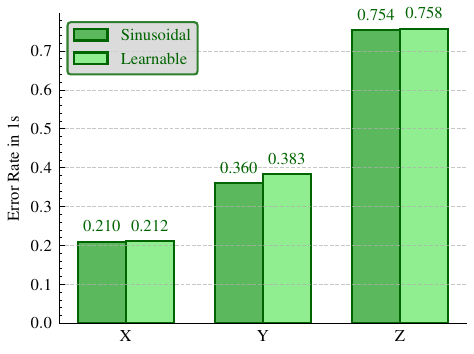}
  \caption{\textbf{Positional Encoding:} Sinusoidal positional encoding achieves better model accuracy than learnable encoding for predicting $\mathbf{X}$, $\mathbf{Y}$, and $\mathbf{Z}$ components of the robot pose.}
  \label{fig:pos_encoding}
  % \vspace{-1.2em}
\end{figure}

\textbf{Positional encoding} is crucial for addressing the permutation equivariance of Transformers, which, by design, lacks inherent sensitivity to input sequence order. This characteristic necessitates the explicit provision of positional information to enable the model to effectively process sequential data. Learnable positional encodings, typically implemented as trainable vectors added to input embeddings, have found favor in CV applications~\cite{he2022masked}. Conversely, non-learnable encodings, such as the sinusoidal functions introduced in the seminal work by~\citet{vaswani2017attention}, have demonstrated efficacy in NLP tasks.
This divergence in methodological preference may stem from inherent differences in the statistical properties of data modalities. CV tasks often involve spatially structured data where absolute positional information may be less critical than relative relationships between local features. In such contexts, learnable encodings may offer greater flexibility in adapting to task-specific positional dependencies. Conversely, NLP tasks frequently rely on precise word order and long-range dependencies, where the fixed nature of non-learnable encodings may provide a beneficial inductive bias~\cite{weng2024navigating}.

To empirically investigate the relative merits of these approaches on robot mobility tasks, we conduct a comparative analysis of learnable positional encodings against sinusoidal encodings as shown in Fig.~\ref{fig:pos_encoding}. Our findings indicate that while both methods achieve comparable asymptotic performance levels, sinusoidal positional encodings exhibit a slight performance advantage.

\begin{figure}[t]
  \centering
  \includegraphics[width=0.7\columnwidth]{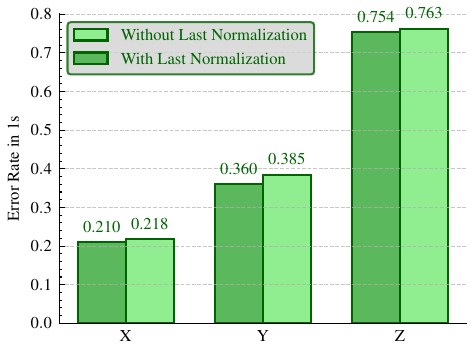}
  \caption{\textbf{Normalizing Output:} Normalizing the \tr~output before passing the embeddings to the task decoder improves model performance.}
  \label{fig:last_layernorm}
  % \vspace{-1.2em}
\end{figure}
\textbf{Normalization layers}, such as LayerNorm~\cite{ba2016layer} or RMSNorm~\cite{zhang2019root}, have been shown to play a crucial role in stabilizing the training of Large Language Models (LLMs)~\cite{loshchilov2024ngpt}. By normalizing the activations of hidden units, these layers help to address issues such as vanishing/exploding gradients and improve the overall stability of the training process~\cite{xiong2020layer}. In this study, we investigate the impact of applying RMSNorm layer immediately before the task head.

Our experiment results, depicted in Fig.~\ref{fig:last_layernorm}, demonstrate an advantage for a model incorporating RMSNorm layer before the task head. This configuration consistently exhibits improved generalization performance and enhanced training stability compared to a model without the final RMSNorm. This finding suggests that normalizing the final embedding vector before passing it to the task head can benefit model performance, potentially by facilitating more effective gradient flow and thus improving the robustness of the model's predictions.

\subsection{Experiment Results from a Robotics Perspective} \label{sec:robotic_perspective}

\begin{figure}[h]
  \centering
  \includegraphics[width=0.7\columnwidth]{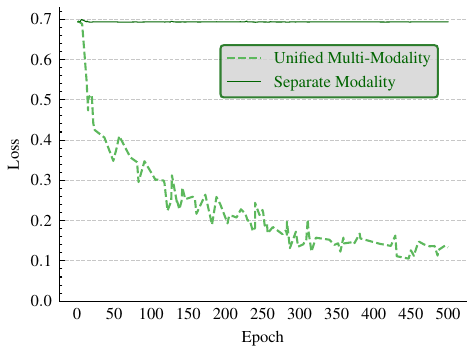}
  \caption{\textbf{Kinodynamics Understanding:} Without unified latent representation the model cannot capture temporal dependencies and understand kinodynamic transitions, resulting in an almost flat learning curve.}
  \label{fig:unified_state}
  % \vspace{-1.2em}
\end{figure}
\textbf{Unified latent space representation} offers a significant advantage in simultaneously addressing FKD, IKD, and BC. This unified approach facilitates a more holistic understanding of the robot's state and its interaction with the environment. To evaluate the efficacy of this unified representation, we perform a targeted ablation study. We train \coder~based on the objectives outlined by~\citet{nazeri2024vertiencoder} and augment them with additional objectives specifically designed to probe the model's capacity of kinodynamics understanding. 

A key component of this ablation involves the introduction of a sequence order prediction objective. This objective aims to assess whether the model can effectively discern the temporal evolution of robot and environment dynamics. During training, the model is presented with input sequences in two configurations: (1) 50\% of the time, the input sequence is presented in its natural temporal order; (2) the remaining 50\% of the time, the input sequence is randomly shuffled, disrupting the temporal coherence. The model is then tasked to classify whether an unseen sequence is presented in its original order or is shuffled, testing the model's ability to capture temporal dependencies and understand kinodynamic transitions.

As illustrated in Fig.~\ref{fig:unified_state}, our findings demonstrate a clear distinction in model performance based on the input representation. When the model is provided with separate, non-unified tokens, it exhibits a limited capacity of understanding the underlying kinodynamics and the learning loss barely drops. This suggests that processing information in a fragmented manner hinders the model's ability to capture temporal relationships and kinodynamic evolution, which is aligned with the findings by~\citet{zhou2024dinowm}. It may be possible to compensate by training with a larger dataset, which, however, is not always available in robotics.

Conversely, the utilization of a unified latent space representation significantly enhances the model's ability to discern temporal order and, consequently, understand the dynamics of the system. By consolidating relevant information into a single, cohesive representation, the model can effectively capture the interdependencies among different modalities and their evolution over time. This highlights the importance of a unified latent space representation in enabling robotic models to effectively learn and reason about complex dynamic systems when trained on limited data, in contrast to NLP and CV tasks where the data acquisition is easier.

\begin{figure}[h]
  \centering
  \includegraphics[width=1\columnwidth]{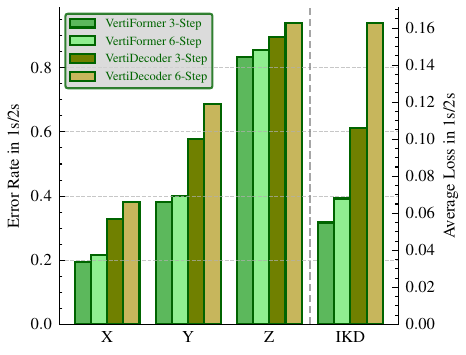}
  \caption{\textbf{Prediction Horizon:} \former~is capable of predicting a longer horizon without losing much accuracy due to its non-autoregressive nature.}
  \label{fig:pred_horizon}
  % \vspace{-1.2em}
\end{figure}
\textbf{Prediction horizon} is a critical factor in navigation planning. While longer prediction horizons can potentially lead to better planning by considering long-term effects, they also introduce greater uncertainty. This is because errors in early predictions can accumulate and lead to significant deviations in subsequent predictions. This issue is particularly relevant for autoregressive models such as the \vertidecoder~part of \former, where each prediction is based on the previous one. In such models, even a small error in the initial steps can propagate and amplify over time, causing the predicted trajectory to drift further away from the true path. To evaluate the impact of prediction horizon, we compare the performance of the autoregressive \vertidecoder~with the non-autoregressive \former, specifically focusing on their ability to maintain accuracy over long horizons. The results, shown in Fig.~\ref{fig:pred_horizon}, demonstrate that \former~is capable of predicting a longer horizon (two seconds) with less drift compared to its autoregressive counterpart even with a shorter horizon (one second). This highlights the advantage of non-autoregressive models in tasks requiring long-term prediction, as they are less susceptible to error accumulation.

\subsection{Experiment Results of Robotic Objective Functions} \label{sec:objective_perspective}
\begin{figure}[h]
  \centering
  \includegraphics[width=\columnwidth]{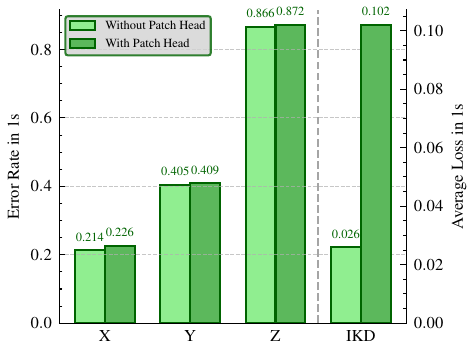}
  \caption{\textbf{Patch Prediction Head:} The inclusion of a patch reconstruction head results in a degradation of overall model performance. This counterintuitive result can be attributed to the inherent difficulty in accurately predicting the detailed structure of off-road terrain topography.}
  \label{fig:patch_head}
  % \vspace{-1.2em}
\end{figure}

\textbf{Patch prediction head}, as an auxiliary head to learn environment kinodynamics, was first introduced by~\citet{nazeri2024vertiencoder}. However, we find that the high complexity of off-road terrain topography and the potential presence of noise or occlusion within the input data create a challenging reconstruction task (see Fig.~\ref{fig:cover}). Consequently, the patch prediction head often generates inaccurate reconstructions, introducing noise into the learning process and negatively impacting the performance of the primary tasks, i.e., FKD, IKD, and BC. This suggests that the auxiliary task of patch reconstruction, in this specific domain, may introduce a conflicting learning signal that hinders the model's ability to effectively learn the desired representations for the main objectives (Fig.~\ref{fig:patch_head}).

\textbf{MM vs NTP vs End-to-End} (End2End) are currently the prominent approaches in CV, NLP, and robotics respectively. However, it is unclear what is the best approach for robot learning, especially learning off-road mobility. We present a comparative analysis of model performance utilizing the MM paradigm within an encoder architecture (\coder, Fig.~\ref{fig:former} left trained alone with MM), a decoder employing autoregressive NTP (\vertidecoder, Fig.~\ref{fig:former} right trained alone without cross-attention), and a non-Transformer-based End2End approach. We then further contrast these approaches with \former,  which adopts a non-autoregressive approach to NTP and MM (Fig.~\ref{fig:former}, trained end-to-end). 

\begin{figure}[t]
  \centering
  \includegraphics[width=\columnwidth]{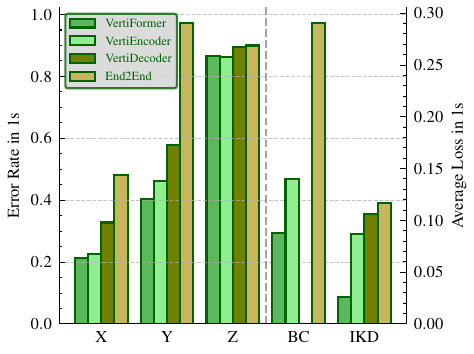}
  \caption{\textbf{MM vs NTP  vs End2End:} \former~achieves best accuracy across FKD, IKD, and BC compared to \coder~(MM), \vertidecoder~(NTP), and End2End.}
  \label{fig:mm_vs_ntp}
  % \vspace{-1.2em}
\end{figure}

To be specific, an encoder model leverages the principles of MM, wherein portions of the input sequence (poses, actions, and terrain patches) are masked, and the model is trained to reconstruct the masked elements. This approach has demonstrated success in capturing contextual dependencies and learning robust representations~\cite{nazeri2024vertiencoder};
A decoder model employs NTP, a prevalent technique in autoregressive sequence generation. In this paradigm, the model predicts the subsequent element in a sequence conditioned on the preceding elements. For both encoder and decoder models, we use the same unified latent space representation presented in Sec.~\ref{sec:unified}. The specialized non-Transformer-based End2End approach uses Resnet-18~\cite{he2015deep} as a patch encoder and fully connected layers as the task heads. While more complex models might offer higher accuracy, we choose ResNet-18 to balance performance with the computational constraints of our robotic platform, making it well-suited for deployment on robots with limited on-board processing capabilities, compared to deeper networks like ResNet-50 or ResNet-101. More information about End2End model architecture is provided in Appendix~\ref{app:architecture}.

As illustrated in Fig.~\ref{fig:mm_vs_ntp}, our findings indicate that \former, a non-autoregressive Transformer, exhibits superior performance across various evaluation metrics, including FKD, IKD, and BC error rates, in the context of one-second prediction horizon. Compared to \vertidecoder, \former~predicts multiple future states simultaneously (i.e., non-autoregressively), which contributes to its better accuracy. These results suggest that the enhanced contextual awareness afforded by the non-autoregressive approach contributes to improved predictive accuracy. Note that \vertidecoder~cannot perform BC directly, as it has access to both action and pose at each step. Unlike \coder~\cite{nazeri2024vertiencoder}, \former~does not train different downstream heads separately each time and all tasks contribute to the performance of each other all together, which results in \former's lowest error rate in most cases (except for $\mathbf{Z}$ prediction). 
Across all kinodynamics tasks, End2End achieves the highest error rate, which shows the benefits of using Transformers for kinodynamic representation and understanding during off-road mobility tasks. 

Beyond the observed performance gains and training stability, \former~demonstrates the capacity of concurrent execution of multiple tasks, not only during training but also during inference. This is particularly relevant in robotics, where real-time control is required and sometimes some modalities may not be available during inference. For example, without a global planner, action sampler, or in the presence of sensor degradation, the robot may not always have access to desired future robot poses, candidate actions, or future terrain patches, respectively. 
Furthermore, the usage of a learned mask within the decoder part of \former~is posited to capture salient distributional characteristics of the data, effectively serving as a condensed representation during inference. This learned representation facilitates adaptation to new tasks where action or pose is missing.

\section{Robot Experiments}
\label{sec:experiments}
\begin{figure}[ht]
    \centering
    \includegraphics[width=0.75\columnwidth]{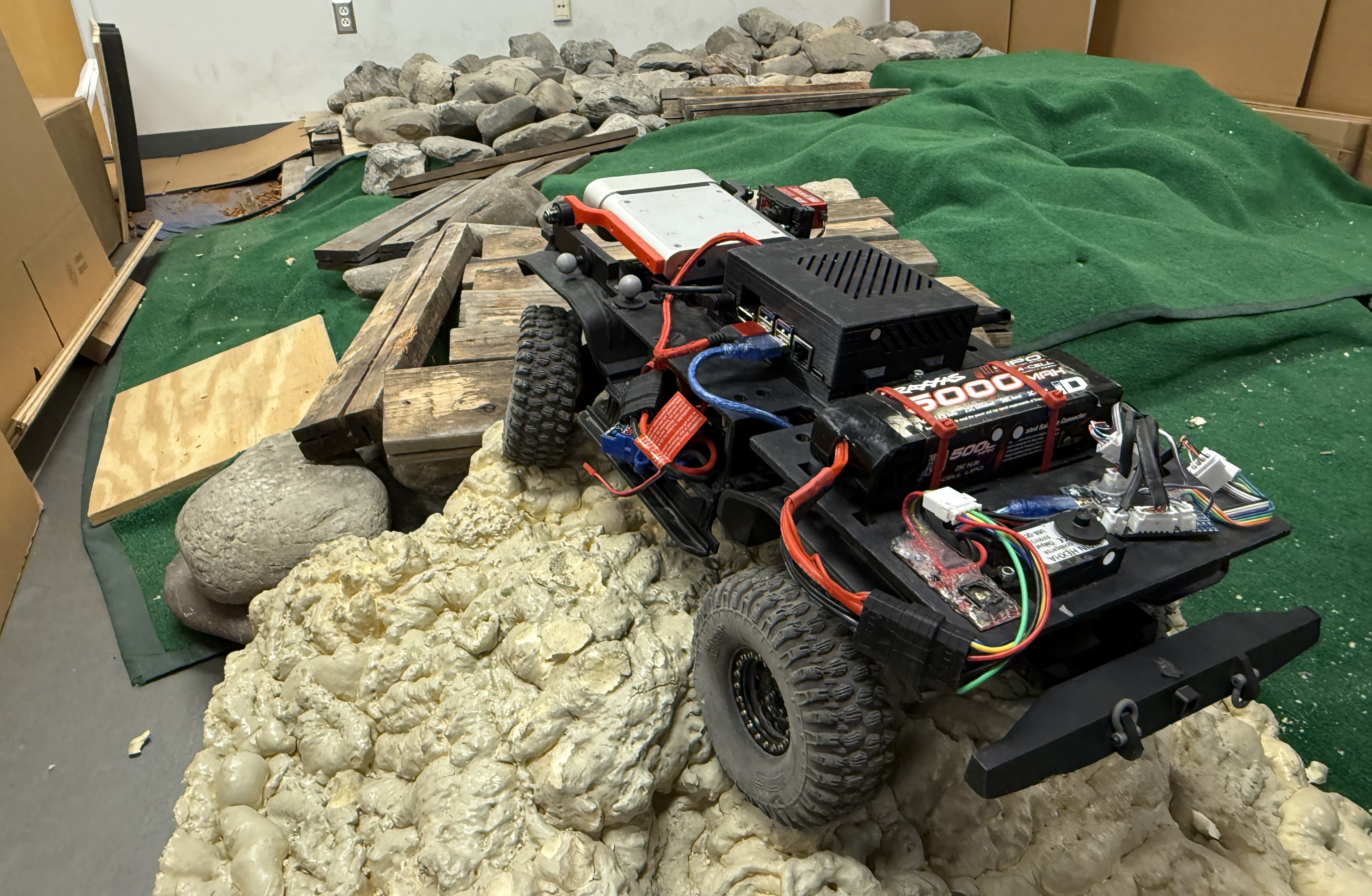}
    \caption{Unseen Test Environments with Rocks/Boulders, Wooden Planks, AstroTurf, and Expanding Foam. }
    \label{fig::test_env}
    \vspace{-1.2em}
\end{figure}

We implement \former's FKD, IKD, and BC on an open-source Verti-4-Wheeler (V4W) ground robot platform. The experiments are carried out on a 4 m $\times$ 2.5 m testbed made of rocks/boulders, wooden planks, AstroTurf with crumpled cardboard boxes underneath, and modular 0.8 m $\times$ 0.75 m expanding foam to represent different types of vertically challenging terrain with different friction coefficients and varying deformability (Fig.~\ref{fig::test_env}). The modular foam and rocks/boulders do not deform, while the rocks may shift positions under the weight of the robot. On the other hand, the wooden planks and AstroTurf are completely deformable and change the terrain topography during wheel-terrain interactions. The one-hour training dataset used (see details of the dataset in Appendix~\ref{app:implementation}) only consists of robot teleoperation on the rigid rock/boulder testbed and hence the experiment testbed is an unseen environment, posing generalization challenges for \former. 

\subsection{Implementation and Metrics}

\subsubsection{FKD} \former's FKD task is integrated with the  MPPI planner~\cite{williams2017model} with 1000 samples and a horizon of 18 steps. We sample across a range of control sequences centered around the last optimal control sequence selected by the robot. The first three actions in a sampled control sequence are passed to \former~along with six past poses, actions, and terrain patches at 3 Hz consisting of one second. The model is repeated six times and outputs 18 future poses of the robot, which are combined to create one candidate trajectory. All 1000 candidate trajectories are then evaluated by a cost function, which calculates the cost of each trajectory based on the Euclidean distance to the goal and roll and pitch angles of the robot. Higher distance, roll, and pitch values are penalized with higher cost. Based on the cost function, MPPI outputs the best control sequence moving the robot forward at 3 Hz. The V4W executes the first action and replans.

\subsubsection{IKD} We integrate \former's IKD task with a global planner based on Dijkstra's algorithm~\cite{dijkstra1959note}, which minimizes traversability cost on a traversability map~\cite{pan2024traverse}. The global planner generates three desired future poses with the lowest cost and passes them to \former, which also has access to six past poses, actions, and terrain patches. \former~then produces three future actions to drive the robot to the three desired future poses. Similarly to FKD, the V4W executes the first action and then replans at 3 Hz. 

\subsubsection{BC} We implement \former's BC by passing in six past poses, actions, and terrain patches to \former. The model outputs three future actions to take. Similarly to FKD and IKD, the first action is executed by V4W and the replanning of BC runs at 3 Hz.

\begin{table*}[h]
    \centering
    % \vspace{0.5em}
    \begin{NiceTabular}{lccccc}

    \toprule
    Task & Model & \textbf{{Success Rate}~$\uparrow$} & \textbf{{Traversal Time}~$\downarrow$} & \textbf{{Mean Roll}~$\downarrow$} & \textbf{{Mean Pitch}~$\downarrow$} \\
    \midrule
    %%%%%%%%%%%%%%%% group by task
    \Block{4-1}{FKD}  & \tal          & 8/10           & $ 11.80         \pm 0.87 $ & $0.198\pm 0.38$          & $ \textbf{0.086} \pm \textbf{0.07}$ \\
                      & \vertidecoder & 6/10           & $ 15.12         \pm 1.78 $ & $0.180 \pm 0.30$       & $0.114 \pm 0.09$  \\
                      & \coder        & \textbf{10}/10 & $\textbf{8.58}  \pm 1.54 $ & $0.189 \pm 0.23$ & $0.116 \pm 0.08$  \\
                      & \former       & \textbf{10}/10 & 9.42           $\pm$ \textbf{0.61} & $\textbf{0.169} \pm \textbf{0.17}$ & $0.096 \pm 0.08$ \\
    \midrule
    \Block{3-1}{IKD}  & \vertidecoder & \textbf{10}/10 &   15.92        $\pm$ \textbf{1.08} & $0.181 \pm 0.23$ & $0.125 \pm 0.08$ \\
                      & \coder        & 7/10           & \textbf{13.99} $\pm$ 3.27 & $\textbf{0.136} \pm 0.14$ & $\textbf{0.069} \pm \textbf{0.07}$\\
                      & \former       & 8/10           & $17.16          \pm 6.10$ & $\textbf{0.136} \pm \textbf{0.10}$ & $0.077 \pm \textbf{0.07}$\\
    \midrule
    \Block{2-1}{BC}   & \coder        & \textbf{9}/10   & 13.49          $\pm$ \textbf{3.33} & $0.175 \pm 0.37$ & $\textbf{0.089} \pm 0.09$\\
                      & \former       & 8/10            & \textbf{12.64} $\pm$ 3.89 & $\textbf{0.154} \pm \textbf{0.11}$ & $0.099 \pm \textbf{0.08}$\\
    
    \bottomrule
    \end{NiceTabular}
    
    \caption{Robot experiments with \former, \coder, \vertidecoder, and \tal.}\label{tab::robot_exp}
    \vspace{-2.0em}
\end{table*}

For FKD and IKD, a trial is deemed successful if the robot reaches the defined goal without rolling over or getting stuck. For BC without explicit goal information, a trial is considered successful if the robot successfully traverses the entire testbed.

\subsection{Results and Discussions}

The results of the three methods are then compared to MPPI using TAL~\cite{datar2024terrainattentive}, a highly accurate forward kinodynamic model specifically designed for vertically challenging terrain. We report the success rate, average traversal time, and mean roll and pitch angles in Table~\ref{tab::robot_exp}.

Our observations reveal a nuanced performance difference between \coder~and \former, particularly concerning BC and IKD. \coder~excels in BC due to its specialized BC task head, a dedicated component trained specifically for this task. This specialized training allows \coder~to effectively leverage the provided data for imitation learning. In contrast, \former~approaches BC in a zero-shot manner. It is not explicitly trained on BC, relying instead on its modality masking strategy. This masking effectively handles missing modalities by replacing them with a trained mask, enabling the model to infer behavior without direct BC training. While this approach allows \former~to perform BC without specialized training, it also explains why \coder, with its dedicated head, achieves a higher success rate. A similar trend is observed with IKD. \coder~benefits from a specialized IKD head, again trained explicitly for this task. And \vertidecoder~has access to both predicted and actual actions and poses at each time step, providing richer guidance for the IKD process. This richer information stream in \vertidecoder~is the reason for achieving a higher success rate, especially considering the inherent difficulty of IKD compared to FKD. \former, however, faces a challenge in IKD and takes longer to finish the traversal.  The masking strategy, while effective for missing modality, is not as accurate as the actual modality.

Regarding FKD, the architectural difference between \former~and \coder~causes different navigation behaviors. \coder's specialized task head for FKD treats each future step independently without any attention weights between steps. While this approach facilitates faster MPPI initial convergence due to a lack of cross attention, it can also lead to drift, causing inconsistencies between predicted steps and ultimately resulting in a larger traversal time standard deviation across trials. While \coder's MPPI converges quickly, it struggles with long-term consistency. \former~takes a different approach. By employing attention and cross-attention mechanisms between historical and future steps, it dynamically incorporates past information into future predictions. This allows \former~to consider the historical context through cross-attention and causal masking when predicting future states, leading to more coherent and consistent predictions. Consequently, although MPPI might require more time to converge on a path with \former, once it does, the resulting behavior is more robust and less variable across trials, reflected in a smaller traversal time standard deviation.  The attention mechanism allows \former~to learn more complex temporal dependencies, which are crucial for accurate long-term prediction in FKD. 

\section{Limitations}
\label{sec:limitations}
Although \former~can capture long-range dependencies through additional context tokens, it requires re-training if we want to change the prediction horizon, while autoregressive models can predict any number of steps into the future without re-training. 
As illustrated in Fig.~\ref{fig:qualitative} of Appendix~\ref{app:qualitative}, our model demonstrates a deficiency in accurately executing a turning maneuver. Such failures stem from long-horizon (1 second), non-autoregressive predictions in one step accentuated by the inaccuracy of terrain reconstruction caused by the high degree of complexity present in off-road topographical formations. This also reflects on the accuracy of predicting $\mathbf{Z}$. A further limitation stems from the use of a mask in place of true modality data. While this approach empowers the model with multi-task capability and to handle missing information, it nonetheless falls short of leveraging the full potential of the actual modalities. 

It is crucial to acknowledge that our observations are primarily associated with the challenges inherent in wheeled locomotion on complex, vertically challenging, off-road terrain and do not necessarily generalize to other robotic domains such as visual navigation or manipulation. In visual navigation, the robot typically relies on visual cues and image processing to perceive its environment and plan its path. In manipulation tasks, the focus is on interacting with objects rather than negotiating through complex terrain. 
Further investigation is required for general visual navigation and manipulation.

\section{Conclusions} 
\label{sec:conclusion}
In this work, we introduce \former, a novel data-efficient multi-task Transformer designed for learning kinodynamic representations on vertically challenging, off-road terrain. \former~demonstrates the capacity to simultaneously address forward kinodynamics learning, inverse kinodynamics learning, and behavior cloning tasks, only using one hour of training data. 
Key contributions include a unified latent space representation enhancing temporal understanding, multi-context tokens enabling multi-step prediction without autoregressive feedback, and a learned masked representation facilitating multiple off-road mobility tasks simultaneously and acting as a proxy for missing modalities during inference. All three contributions improve robustness and generalization of \former~to out-of-distribution environments.
We provide extensive experiment results and empirical guidelines for training Transformers under extreme data scarcity. 
Our evaluations across all three downstream tasks demonstrate that \former~outperforms baseline models, including \tal~\cite{datar2024terrainattentive}, \coder~\cite{nazeri2024vertiencoder}, \vertidecoder, and end-to-end approaches, while exhibiting reduced overfitting and improved generalization and highlighting the efficacy of the proposed architecture and training methodology for learning kinodynamic representations in data-constrained settings. Physical experiments also demonstrate that \former~can enable superior off-road robot mobility on vertically challenging terrain.

\section*{Acknowledgments}
This work has taken place in the RobotiXX Laboratory at George Mason University. RobotiXX research is supported by National Science Foundation (NSF, 2350352), Army Research Laboratory (ARL, W911NF2220242, W911NF2320004, W911NF2420027, W911NF2520011), Air Force Research Laboratory (AFRL) and US Air Forces Central (AFCENT, GS00Q14OADU309), Google DeepMind (GDM), Clearpath Robotics, and Raytheon Technologies (RTX).

%% Use plainnat to work nicely with natbib. 
% \clearpage
\bibliographystyle{plainnat}
\bibliography{bibliography}

\clearpage
\appendices % or \appendix if there's only one appendix else \appendices

\section{Model Architecture} \label{app:architecture} % model architecture
\begin{table}[h]
\centering
\caption{\former~Architecture Parameters.}
\begin{tabular}{ll}
\toprule
\multicolumn{2}{c}{\coder} \\
\midrule
Layers & 6 \\
Normalization & RMSNorm~\cite{zhang2019root} \\
Hidden size $D$ & 512 \\
Heads & 8 \\
MLP size & 512 \\
Dropout & 0.3 \\
Activation & GELU~\cite{hendrycks2017bridging} \\
Pre-Norm & True \\
PositionalEncoding & Sinusoidal \\
\midrule
\multicolumn{2}{c}{\vertidecoder} \\
\midrule
Layers & 4 \\
Normalization & RMSNorm~\cite{zhang2019root} \\
Hidden size $D$ & 512 \\
Heads & 8 \\
MLP size & 512 \\
Dropout & 0.3 \\
Activation & GELU~\cite{hendrycks2017bridging} \\
Pre-Norm & True \\
PositionalEncoding & Sinusoidal \\
\bottomrule
\end{tabular}
\label{tab:transformer_params}
\end{table}

\begin{table}[h]
\centering
\caption{End2End Architecture Parameters.}
\begin{tabular}{ll}
\toprule
\multicolumn{2}{c}{End2End} \\
\midrule
Patch Encoder & Resnet-18 \\
Normalization & batch norm~\cite{sennrich2016neural} \\
Hidden Layer 1& 256 \\
Hidden Layer 2& 512 \\
Hidden Layer 3& 64 \\
Activation & Tanh \\
Dropout & 0.2 \\
\bottomrule
\end{tabular}
\label{tab:e2e_params}
\end{table}

\section{Implementation Details} \label{app:implementation}
% We use an open-source Verti-4-Wheeler (V4W) platform, as described by~\citet{Datar2024a} as our robot platform. The V4W is equipped with a Microsoft Azure Kinect RGB-D camera and an NVIDIA Jetson Xavier processor. On the software side, \former~is implemented with PyTorch and trained on a single A5000 GPU with 24GB memory while only occupying 2GB of memory.

We use an open-source V4W robotic platform, as detailed by~\citet{datar2024wheeled}, for physical evaluation. The V4W platform is equipped with a Microsoft Azure Kinect RGB-D camera to build elevation maps~\cite{miki2022elevation} and an NVIDIA Jetson Xavier processor for onboard computation. The proposed \former~model is implemented using PyTorch and trained on a single NVIDIA A5000 GPU with 24GB of memory, demonstrating efficient memory utilization with a peak memory footprint of only 2GB.

\noindent\textbf{Optimization:} we use the AdamW optimizer~\cite{loshchilov2019decoupled} with learning rate of $5e^{-4}$ and weight decay of $0.08$. We train \former~for 200 epochs with a batch size of 512.

% \textbf{Dataset:} We utilize the dataset introduced in \tal~\cite{datar2024terrainattentive}, collected on a 3.1 m $\times$ 1.3 m rock testbed with a maximum height of 0.6 m. 
% The dataset consists of 30 minutes of data on a planar surface and 30 minutes on the rock testbed. The dataset has a variety of 6-DoF vehicle states including vehicle rollover and getting stuck achieved while manually teleoperating the robot over the rock testbed during data collection. This variety is achieved because of the modularity of the rock testbed enabling flexible reconfiguration for data collection and mobility experiments. The dataset contains VIO for vehicle state estimation, elevation maps built from depth images, and teleoperated vehicle controls including throttle and steering commands.

\noindent\textbf{Dataset:} We utilize the dataset introduced by \tal~\cite{datar2024terrainattentive}, which was collected on a 3.1 m $\times$ 1.3 m modular rock testbed with a maximum height of 0.6 m. The dataset includes 30 minutes of data from both a planar surface and the rock testbed, capturing a diverse range of 6-DoF vehicle states. These states encompass scenarios such as vehicle rollovers and instances of the vehicle getting stuck, all recorded during manual teleoperation over the reconfigurable rock testbed. 
% This modularity allows for a flexible setup, facilitating comprehensive data collection and mobility experiments. 
The dataset comprises visual-inertial odometry for vehicle state estimation, elevation maps derived from depth images, and teleoperation control data, including throttle and steering commands, to provide a holistic view of vehicle dynamics.

\section{Qualitative Results} \label{app:qualitative} % qualitative success and failures
\begin{figure*}[h]
    \centering
    \begin{subfigure}[t]{0.5\textwidth}
        \centering
        \includegraphics[width=\columnwidth]{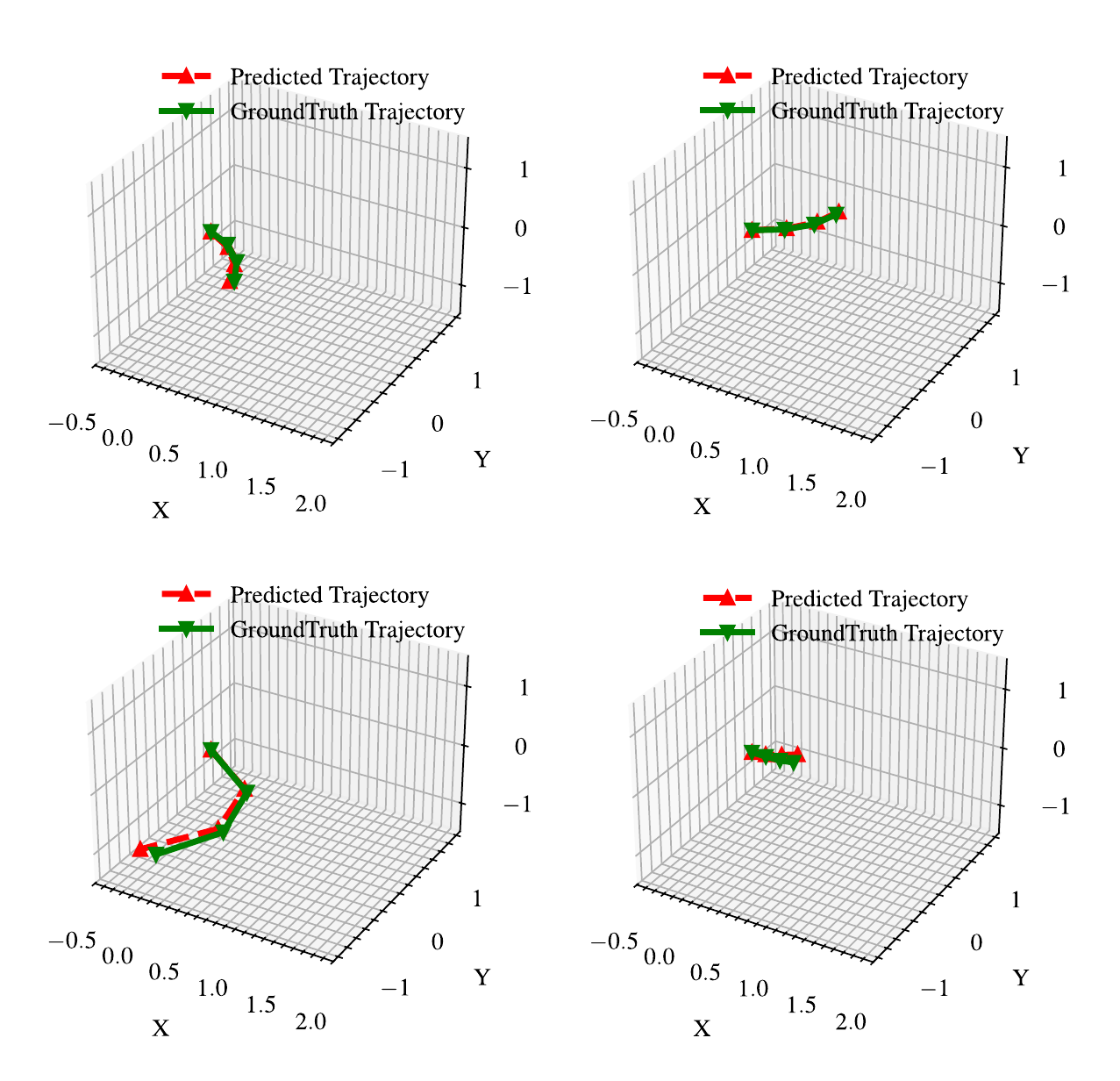}
        \caption{Successful 3-step predictions.}
    \end{subfigure}%
    ~ 
    \begin{subfigure}[t]{0.5\textwidth}
        \centering
        \includegraphics[width=\columnwidth]{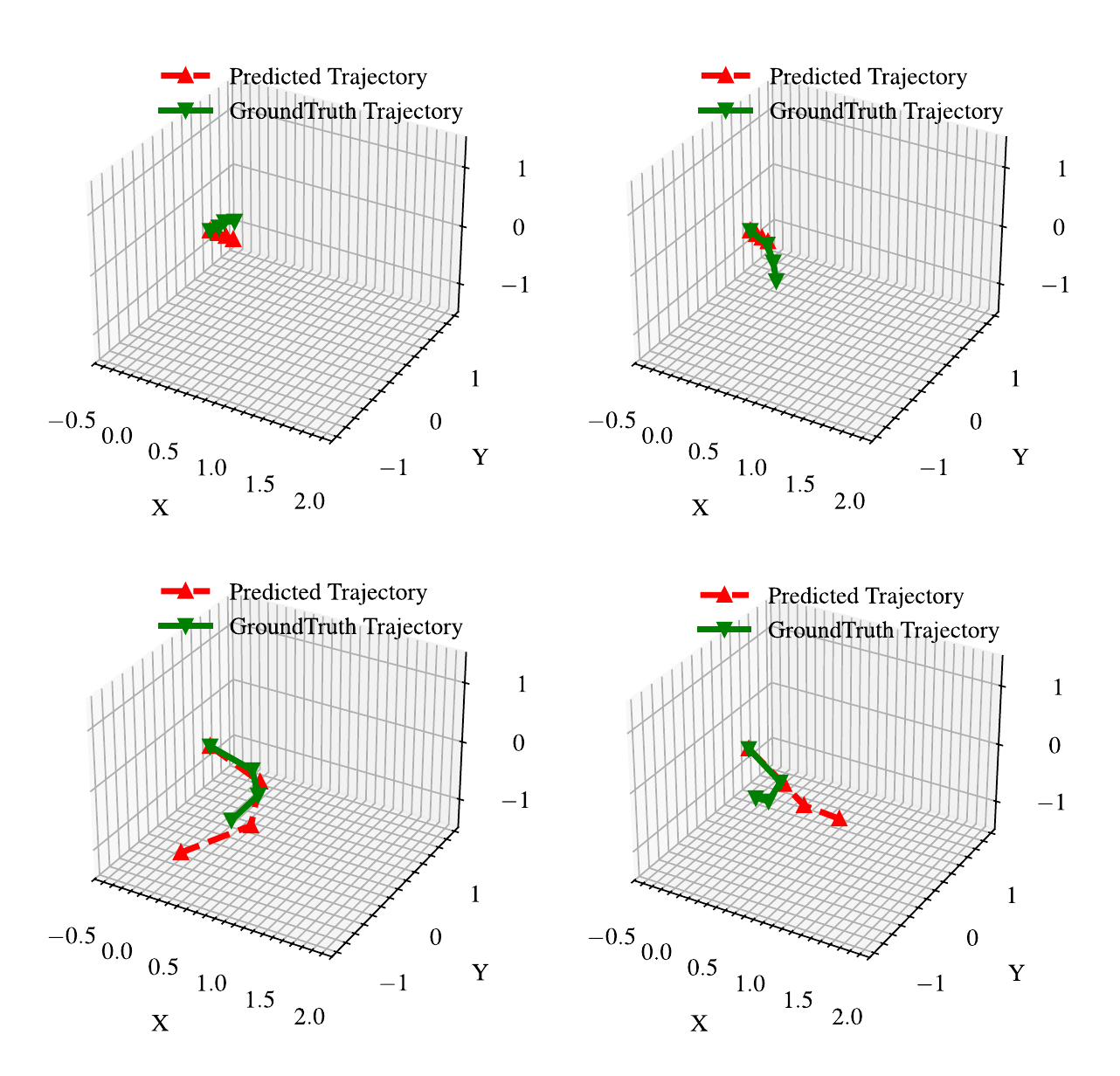}
        \caption{Failed 3-step predictions.}
    \end{subfigure}

    \begin{subfigure}[t]{0.5\textwidth}
        \centering
        \includegraphics[width=\columnwidth]{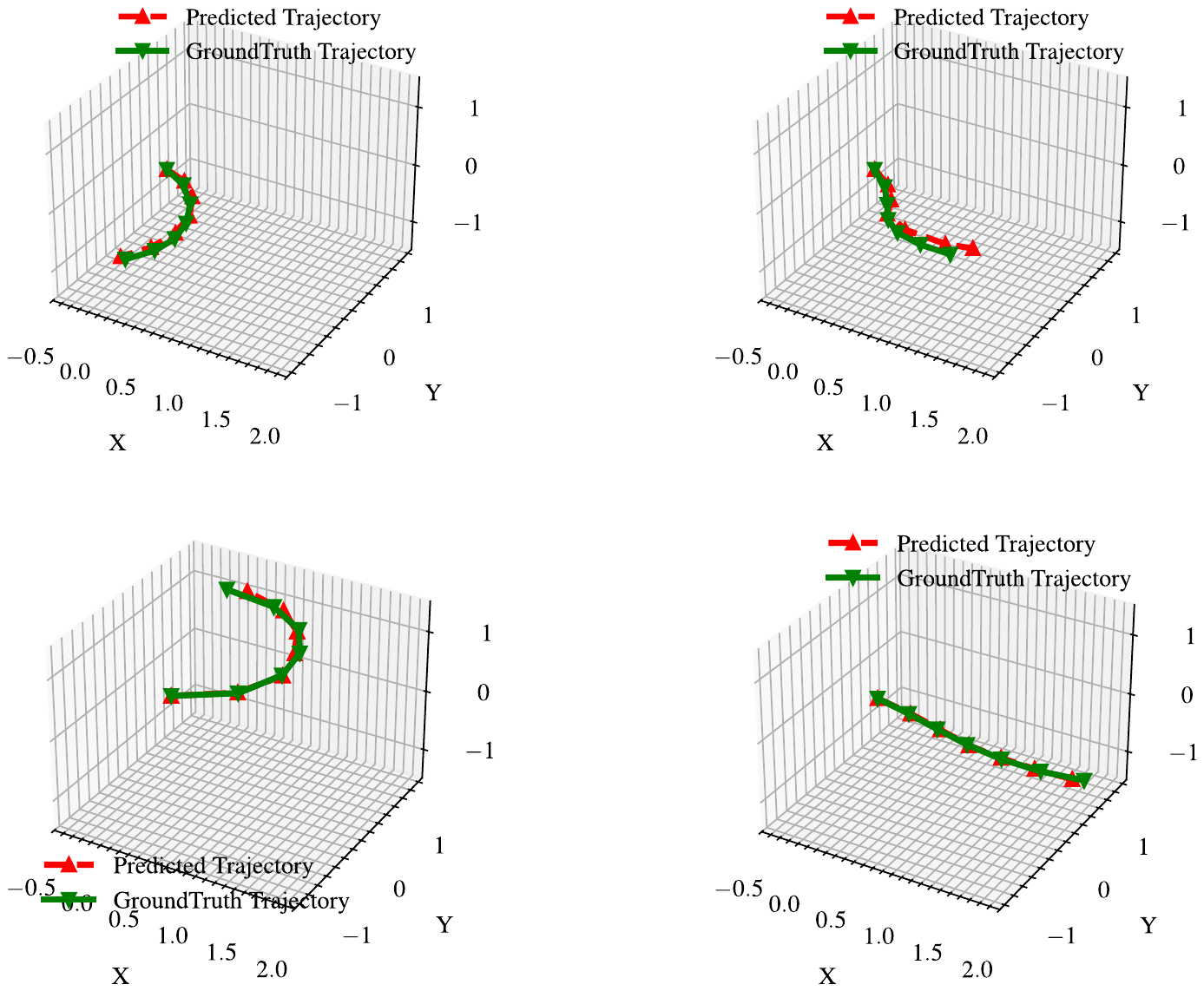}
        \caption{Successful 6-step predictions.}
    \end{subfigure}%
    ~ 
    \begin{subfigure}[t]{0.5\textwidth}
        \centering
        \includegraphics[width=\columnwidth]{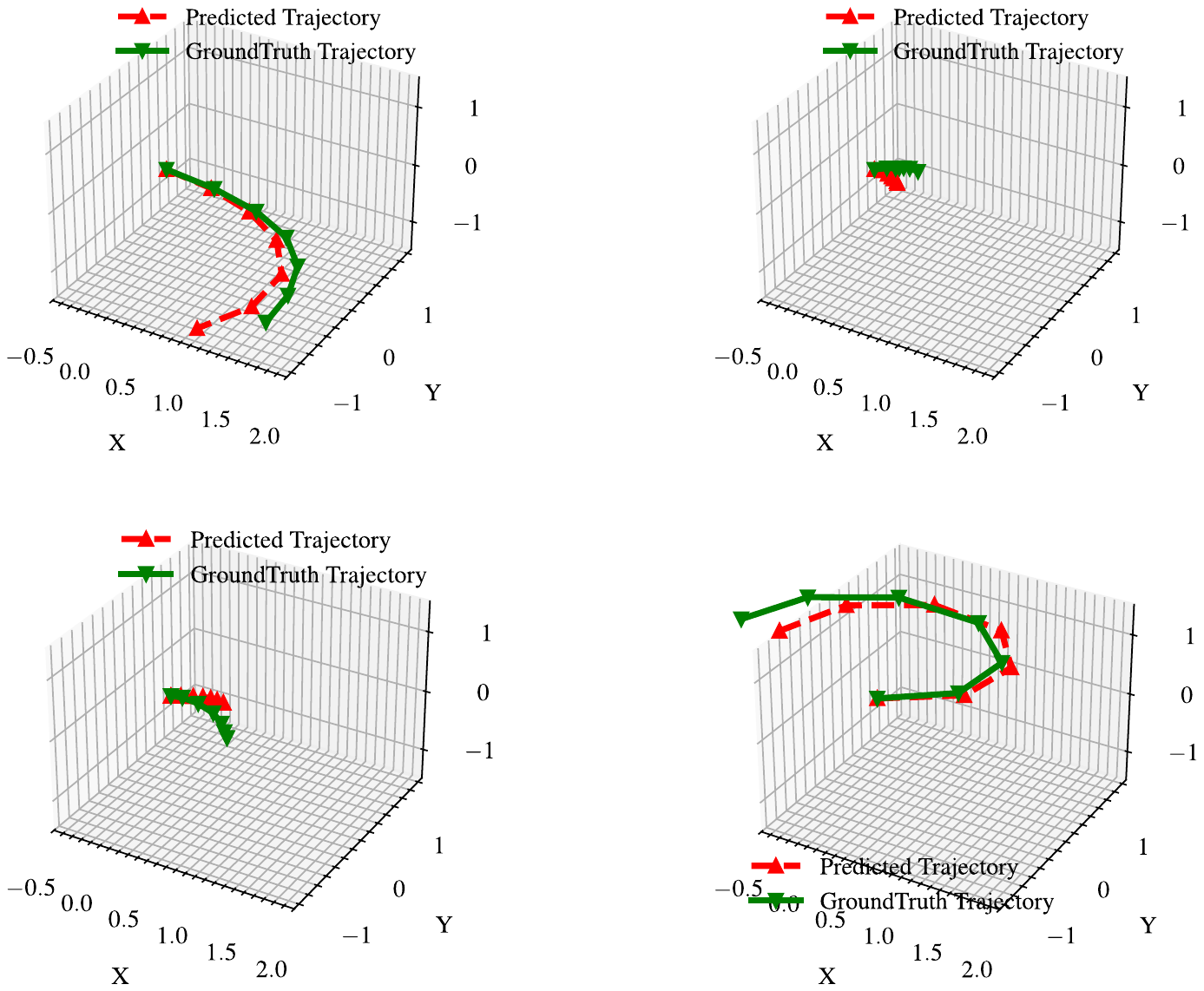}
        \caption{Failed 6-step predictions.}
    \end{subfigure}
    
    \caption{Qualitative Results of 3-Step and 6-Step Successful and Failed Trajectory Prediction over One and Two Second(s).}\label{fig:qualitative}
\end{figure*}

\begin{figure*}[h]
    \centering
    \begin{subfigure}[t]{0.5\textwidth}
        \centering
        \includegraphics[width=\columnwidth]{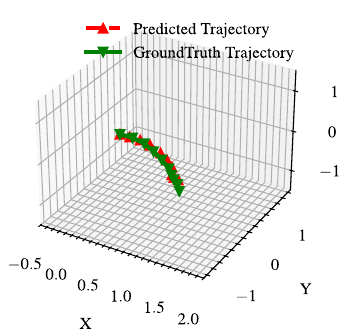}
        \caption{\former~maintains accuracy for longer horizons due to non-autoregressive predictions.}
    \end{subfigure}%
    ~ 
    \begin{subfigure}[t]{0.5\textwidth}
        \centering
        \includegraphics[width=\columnwidth]{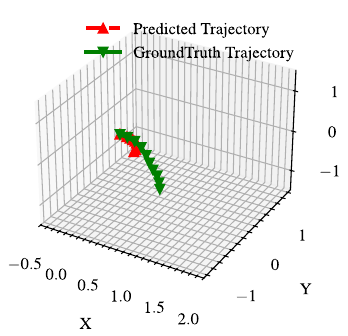}
        \caption{\vertidecoder~drifts from the ground truth due to accumulation of error in autoregressive predictions.}
    \end{subfigure}
    \caption{Qualitative Comparison of Drifting between Non-Autoregressive \former~and Autoregressive \vertidecoder.}\label{fig:drifting}
\end{figure*}

\begin{figure*}[h]
    \centering
    \begin{subfigure}[t]{0.5\textwidth}
        \centering
        \includegraphics[width=\columnwidth]{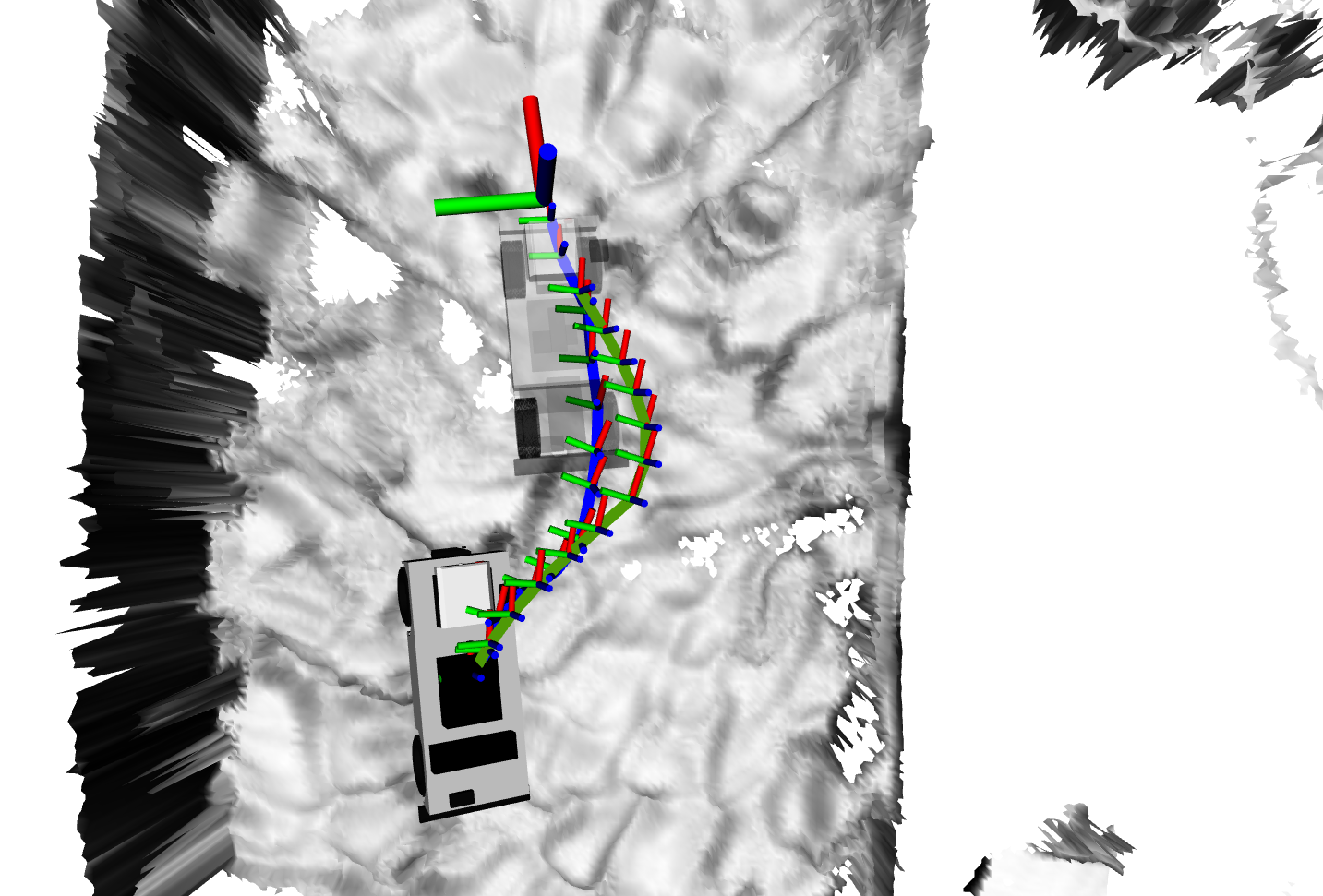}
        % \caption{\former~maintains the accuracy for longer horizons due to non-autoregressive predictions.}
    \end{subfigure}%
    ~ 
    \begin{subfigure}[t]{0.5\textwidth}
        \centering
        \includegraphics[width=\columnwidth]{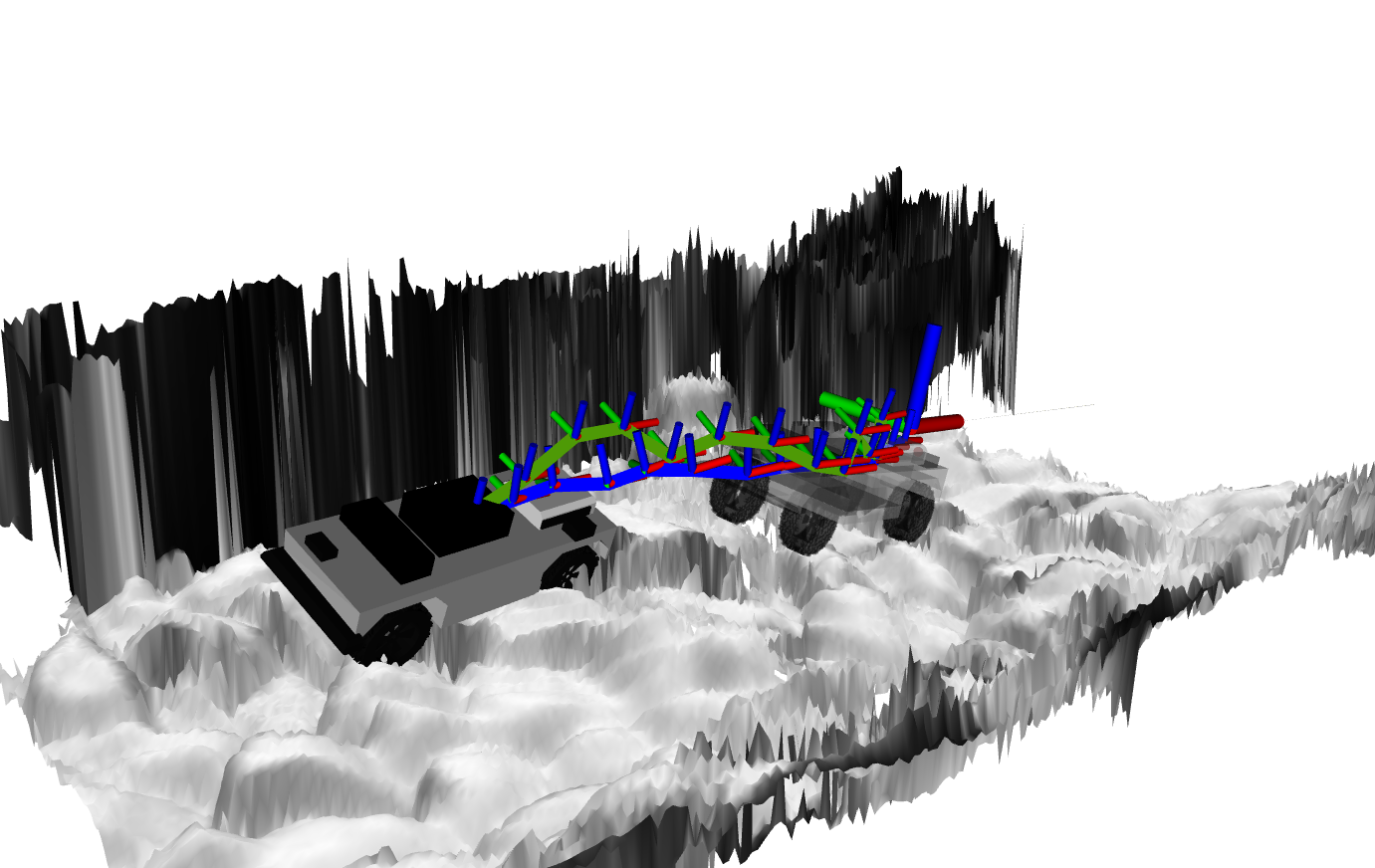}
        % \caption{\vertidecoder~drifts from the ground truth due to accumulation of error in autoregressive predictions..}
    \end{subfigure}
    \caption{Visualization of \former~Predictions in \textcolor{green}{green} and Ground Truth in \textcolor{blue}{blue}.}\label{fig:drifting}
\end{figure*}

\end{document}